\documentclass[11pt,a4paper]{article}
\usepackage[hyperref]{EMNLP/emnlp2020}
\usepackage{times}
\usepackage{latexsym}

\usepackage{graphicx}
\usepackage{tabularx}
\usepackage{soul}

\usepackage{epstopdf}
\usepackage[utf8]{inputenc}

\usepackage{bookmark}
\usepackage{xstring}
\usepackage{url}
\usepackage{multirow}
\usepackage{CJKutf8}

\usepackage{microtype}

\aclfinalcopy 

\newcommand*{\enja}{EN$\rightarrow$JA}
\newcommand*{\jaen}{JA$\rightarrow$EN}

\title{Document-aligned Japanese-English Conversation Parallel Corpus}

\author{Mat{\=\i}ss Rikters, Ryokan Ri, Tong Li and Toshiaki Nakazawa \\
  The University of Tokyo \\
  7-3-1 Hongo, Bunkyo-ku, Tokyo, Japan \\
  \texttt{\{matiss, li0123, litong, nakazawa\}@logos.t.u-tokyo.ac.jp} \\}

\date{}

\begin{document}
\maketitle
\begin{abstract}
Sentence-level (SL) machine translation (MT) has reached acceptable quality for many high-resourced languages, but not document-level (DL) MT, which is difficult to 1) train with little amount of DL data; and 2) evaluate, as the main methods and data sets focus on SL evaluation. To address the first issue, we present a document-aligned Japanese-English conversation corpus, including balanced, high-quality business conversation data for tuning and testing. As for the second issue, we manually identify the main areas where SL MT fails to produce adequate translations in lack of context. We then create an evaluation set where these phenomena are annotated to alleviate automatic evaluation of DL systems. We train MT models using our corpus to demonstrate how using context leads to improvements.
\end{abstract}

\section{Introduction}
\label{sec:intro}

The quality of machine translation (MT) for written text and monologue has vastly improved due to the increased amount of available parallel corpora and recent neural network technologies. However, there is much room for improvement in the context of dialogue or conversation translation. One typical case is the translation from a pro-drop language to a non-pro-drop language where correct pronouns must be supplemented according to the context. The omission of the pronouns occurs more frequently in spoken language than written language. Recently, context-aware MT models attract attention from many researchers \cite{tiedemann-scherrer-2017-neural,voita-etal-2019-good} to solve this kind of problem, however, there are almost no parallel conversation corpora with context information except the rather noisy Open Subtitles corpus \cite{tiedemann-2016-finding}.

A document and sentence-aligned conversation parallel corpus should be advantageous to push MT research in this field to the next stage. In this paper, we introduce a newly constructed document-aligned (DA) Japanese-English conversation corpus, which contains three sub-corpora: Business Scene Dialogue (BSD \cite{rikters-etal-2019-designing}), Japanese translation of AMI Meeting Corpus (AMI \cite{mccowan2005ami}) and Japanese translation of OntoNotes 5.0 (ON \cite{ontonotes}). The corpus contains multi-person conversations in various situations: business scenes, meetings under specific themes, broadcast conversations and telephone conversations.

We supplement the original BSD part with additional data, increasing its size by almost three times. We also enrich the corpus with speaker information and other useful meta-data, and separate balanced versions of development and evaluation data sets.


\section{Related Work}
\label{sec:related}

There are many ready-to-use parallel corpora for training MT systems, but most of them are in written languages such as web crawl, patents \cite{ntcir9-patmt}, scientific papers \cite{NAKAZAWA16.621}. Even though some parallel corpora are in spoken language, they are mostly monologues \cite{cettoloEtAl:EAMT2012,mustc19} or contain a lot of noise \cite{tiedemann-2016-finding,PRYZANT18.30}. Most of the MT evaluation campaigns such as WMT\footnote{\url{http://www.statmt.org/wmt20/}}, WAT\footnote{\url{http://lotus.kuee.kyoto-u.ac.jp/WAT/}} adopt the written language, monologue or noisy dialogue parallel corpora for their translation tasks. Among them, there is only one clean, dialogue parallel corpus \cite{salesky2018slt} adopted by IWSLT\footnote{\url{http://workshop2019.iwslt.org}} in the conversational speech translation task.

JParaCrawl \cite{morishita2019jparacrawl} is a recently announced large English-Japanese parallel corpus built by crawling the web and aligning parallel sentences. Its size is impressive, but it is composed of noisy web-crawled data and has many duplicate sentences. Compared to our corpus, JParaCrawl does not have meta-information and is not DA. 

\newcite{voita-etal-2019-good} evaluate what modern MT systems struggle with when translating from English into Russian and construct new development and evaluation sets based on human evaluation. The sets target linguistic phenomena - dexis, ellipsis and lexical cohesion. The authors also provide code for a context-aware NMT toolkit that improves upon translating these phenomena. In contrast, our development/evaluation sets contain complete documents of consecutive sentences, not broken up into only the sentences requiring context.

\section{Corpus Description}
\label{sec:about}

Our corpus consists of 3 sub-corpora, each of which originates from different sources - BSD, AMI, and ON. BSD was newly constructed, while AMI and ON are translations of the existing English versions of these corpora. Detailed statistics of the sub-corpora are provided in Tables \ref{table:bsd-overview} and \ref{table:ami-on-overview}. BSD consists of the scenes mentioned in Table \ref{table:bsd-overview}, ON has only two different scenes - broadcast conversation and telephone conversation, and all documents from AMI belong to the meeting scene. There is no particular taxonomy associated with these scenes. Word counts for the English side of the sub-corpora are shown in Table \ref{tab:corpora-word-count}. We do not include word counts for the Japanese side since it uses very little spaces and the final word count depends on tokenisation.

\subsection{Construction Process}

\textbf{Business Scene Dialogue}

This sub-corpus was entirely newly created without using any pre-existing resources. We asked professional scenario writers to write monolingual scenarios (documents), and then asked professional translators to translate the documents. This process was done for both En $\leftrightarrow$ Ja directions to ensure a wide range of lexicons and expressions from both languages.

\begin{CJK}{UTF8}{ipxm}
In conversations, the utterances are often very short and vague, therefore it is possible that they should be translated differently depending on the situations where the conversations are taking place. For example, the Japanese expression 「すみません」 can be translated into several English expressions, such as ``Excuse me'', ``Thank you.'' or ``I'm sorry.'', depending on context. By using scene information, it is possible to discriminate the translations, which is hard to do with only the contextual sentences. Furthermore, it may be possible to connect scene information to multi-modal MT, i.e., estimating the scene from visual information. Language used in meetings and presentations is often more formal than general chatting or phone calls. This is especially prevalent in Japanese, which has three distinct levels of politeness in the spoken language. Knowing the scene may be useful for adjusting politeness and formality.
\end{CJK}
\\\\
\textbf{AMI Meeting Parallel Corpus}

The original AMI Meeting Corpus is a multi-modal dataset containing 100 hours of meeting recordings in English.
The parallel version was constructed by asking professional translators to translate utterances from the original corpus into Japanese. Since the original corpus consists of speech transcripts, the English sentences contain a lot of short utterances ({\it e.g., ``Yeah'', ``Okay''}) or fillers ({\it e.g., ``Um''}), and these are translated into Japanese as well. Therefore, the AMI sub-corpus contains many duplicates (see Table \ref{tab:total-unique-other}).
\\\\
\textbf{OntoNotes 5.0}

The original OntoNotes is comprised of various genres of text (news, telephone speech, weblogs, newsgroups, broadcast, talk shows) in three languages (English, Chinese, and Arabic) with additional annotated information - syntax and predicate argument structure, word sense linked to an ontology and coreference. We extracted the English subsets of broadcast conversation (BC) and telephone conversation (Tele), and had professional translators translate them into Japanese.
\\\\
\textbf{Development and Evaluation Sets}

We provide balanced development and evaluation splits from only the BSD sub-corpus as it is the least noisy part. The documents in these sets are balanced in terms of scenes and original languages. The complete statistics are shown in Table \ref{table:devel-eval-overview}.

\begin{table}[t]
    \centering
    \begin{small}
    \begin{tabular}{|l|r|r|r|r|}
    \hline
     & \multicolumn{2}{@{}c@{}|}{\textbf{\jaen}} & \multicolumn{2}{@{}c@{}|}{\textbf{\enja}} \\ \hline 
    \textbf{Scene} & {\scriptsize \textbf{Doc.}} & {\scriptsize \textbf{Sent.}} & {\scriptsize \textbf{Doc.}} & {\scriptsize \textbf{Sent.}} \\ \hline 
        face-to-face & 535 & 16,481 & 458 & 14,858 \\
        phone call & 279 & 8,720 & 256 & 7,770 \\
        general chatting & 233 & 7,674 & 239 & 7,372 \\
        meeting & 224 & 7,647 & 265 & 8,952 \\
        training & 37 & 1,379 & 47 & 1,549 \\
        presentation & 17 & 499 & 53 & 1,899 \\ \hline
        sum & 1,325 & 42,400 & 1,318 & 42,400 \\ \hline
    \end{tabular}
    \end{small}
    \caption{Document (Doc.) and sentence (Sent.) statistics for the full BSD corpus. \jaen\ represents documents written in Japanese and translated into English. \enja\ represents the opposite documents.}
    \label{table:bsd-overview}
\end{table}

\begin{table}[t]
    \centering
    \begin{small}
    \begin{tabular}{|l|c|r|c|c|} 
         \hline
         \textbf{Set (Scene)} & \textbf{Documents} & \textbf{Sentences} & \textbf{PA} & \textbf{WK} \\ 
         \hline
         AMI   & 171  & 110,483  & 4 & 0 \\
         \hline
         ON (BC)        & 27   & 14,354  & 5 & 3 \\
         ON (Tele)      & 46   & 14,075  & 6 & 0  \\
         \hline
    \end{tabular}
    \end{small}
    \caption{Statistics for translated version of AMI and ON corpora and errors detected in \enja \ MT.}
    \label{table:ami-on-overview}
\end{table}

\begin{table}[t]
    \centering
    \begin{tabular}{|l|r|}
    \hline
    & \textbf{Word Count} \\ \hline
    Development & 19,229 \\ \hline
    Evaluation  & 19,619 \\ \hline
    BSD & 750,167 \\ \hline
    AMI & 977,467 \\ \hline
    ON  & 279,709 \\ \hline
    \end{tabular}
    \caption{English side word counts for each of the sub-corpora and development/evaluation sets.}
    \label{tab:corpora-word-count}
\end{table}

\begin{table*}[t]
    \centering
    \begin{tabular}{|l|c|c|c|c|c|c|c|c|c|}
    \hline
        & \multicolumn{4}{c|}{\textbf{Development}} & \multicolumn{4}{c|}{\textbf{Evaluation}} \\ \hline 
     & \multicolumn{2}{@{}c@{}|}{\textbf{\jaen}} & \multicolumn{2}{@{}c@{}|}{\textbf{\enja}} & \multicolumn{2}{@{}c@{}|}{\textbf{\jaen}} & \multicolumn{2}{@{}c@{}|}{\textbf{\enja}}\\ \hline 
    \textbf{Scene} & {\textbf{Doc.}} & {\textbf{Sent.}} & {\textbf{Doc.}} & {\textbf{Sent.}} & {\textbf{Doc.}} & {\textbf{Sent.}} & {\textbf{Doc.}} & {\textbf{Sent.}} \\ \hline 
        face-to-face & 11 & 319 & 12 & 314 & 12 & 381 & 11 & 345 \\
        phone call & 6 & 176 & 7 & 185 & 6 & 163 & 7 & 212 \\
        general chatting & 7 & 223 & 8 & 248 & 7 & 211 & 8 & 212 \\
        meeting & 7 & 240 & 7 & 219 & 7 & 228 & 7 & 229 \\
        training & 1 & 40 & 1 & 23 & 1 & 38 & 1 & 30 \\
        presentation & 1 & 31 & 1 & 33 & 1 & 31 & 1 & 40 \\ \hline
        sum & 33 & 1029 & 36 & 1029  & 34 & 1052 & 35 & 1052 \\ \hline

    \end{tabular}
    \caption{Document (Doc.) and sentence (Sent.) statistics for development and evaluation sets.}
    \label{table:devel-eval-overview}
\end{table*}

\subsection{Analysis}

We extend the analysis conducted for BSD \cite{rikters-etal-2019-designing} to AMI and ON by investigating contextual information requirements for \enja \ MT.
We randomly sample 200 and 100 sentence pairs from ON and AMI respectively. In the case of ON, 50\% of the pairs are from BC and 50\% are from Tele. We translate the sentences with Google Translate\footnote{https://translate.google.com/ (November 2019)} and check the translations for errors, ignoring fluency or minor grammatical mistakes. \\
Unlike the \jaen \ results for BSD, where more than 50\% of errors were due to zero anaphora, there are mainly two types of causes for errors we detected in this analysis - phrase ambiguity (PA) and absence of world knowledge (WK). Most of the errors (Table \ref{table:ami-on-overview}) are caused by PA, for which taking context sentences into account can be considered as a possible solution. On the other hand, the documents in ON-BC contain a variety of named entities (e.g., Shia - one of the two main branches of Islam) and abbreviations (e.g., CPC - Communist Party of China). To solve this, either domain-specific training data or additional mechanisms that take WK into account would be required.

\subsection{Release and Licensing}

The current version of BSD is published on GitHub\footnote{\url{https://github.com/tsuruoka-lab/BSD}} under the  Creative Commons Attribution-NonCommercial-ShareAlike 4.0 International (CC BY-NC-SA 4.0) license. The English OntoNotes is under the LDC User Agreement for Non-Members and the AMI Meeting parallel corpus is published on GitHub\footnote{\url{https://github.com/tsuruoka-lab/AMI-Meeting-Parallel-Corpus}} under Creative Commons Attribution 4.0 license (CC BY 4.0). We plan to release the extended BSD and translations of AMI under the same licenses and are currently negotiating a licensing agreement for the Japanese translations of OntoNotes.

\section{Machine Translation Experiments}
\label{sec:mt-exp}

The conversation corpus alone is not big enough to train real-world NMT systems (as demonstrated by \newcite{rikters-etal-2019-designing}). However, by increasing the size of the high-quality BSD corpus, we managed to train reasonable NMT systems. The full statistics of our data are shown in Table \ref{tab:total-unique-other}.

\subsection{Experiment Setup}
For the SL systems, we used Sockeye \cite{Sockeye:17} to train transformer architecture \cite{NIPS2017_7181} models with the \textit{transformer-base} parameters until convergence on development data (no improvement on validation perplexity for 10 checkpoints). Each model was trained 3 times on a single Nvidia TITAN V (12GB) GPU. The reported BLEU score results are an average of 3 runs. Training time was about 2 days for models with only our data and about 5 days when using WMT data.

To train our context-aware systems, we experimented with two approaches - sentence concatenation \cite{tiedemann-scherrer-2017-neural} with source side factors \cite{sennrich2016linguistic-wmt} and context-aware decoder (CADec \cite{voita-etal-2019-good}). We use the same toolkit and similar parameters as in our SL systems for the former and the CADec toolkit with the default parameters for the latter. For the concatenation context-aware MT, we experimented with two approaches: 1) prepending the previous sentence from the same document, followed by a beginning of sentence tag \textit{$<$bos$>$}, to the source sentence; 2) in addition, providing source side factors to specify if a token represents context or the source sentence.

The source side factors that we used for training were either C or S, representing context and the actual source sentence respectively. Examples of source sentences with context and factors are shown in Table \ref{tab:factors}. The first sentence in the table has no previous context, as it is the first one in the respective document. The second sentence has the first one as context, followed by a beginning of sentence tag \textit{$<$bos$>$}, and so on.

\begin{CJK}{UTF8}{ipxm}
\begin{table}[hb]
    \centering
    \begin{tabular}{|l|}
    \hline
    \textbf{Source sentences} \\ \hline
    \textless{}bos\textgreater{}▁ はい 、 G 社 お客様 相 談 室 の \\ケ イ ト です 。 \\ \hline
    ▁はい 、 G 社 お客様 相 談 室 の ケ イ ト \\ です 。\textless{}bos\textgreater{}▁ ご 用 件 は ? \\ \hline
    ▁ ご 用 件 は ? \textless{}bos\textgreater{}▁ もし もし 、 森 と \\い います 。 \\ \hline
    \textbf{Source side factors} \\ \hline
    C S S S S S S S S S S S S S S \\ \hline
    C C C C C C C C C C C C C C C S S S S S \\ \hline
    C C C C C C C S S S S S S S S \\ \hline
    \end{tabular}
    \caption{Examples of training data source sentences and the respective source side factors for the concatenated context-aware experiments.}
    \label{tab:factors}
\end{table}
\end{CJK}

\subsection{Results}
\label{sec:mt-results}
The results in Table \ref{tab:mt-results} show that decent quality MT models can be trained by using only our corpus (Baseline). For \jaen \ the scores slightly improve by training contextual models (Concatenated and Concatenated + factors), which indicates that there are context-dependent sentences in our evaluation set that benefit from the additional information. We investigate this further by performing human evaluation in Section \ref{sec:human-eval}. We did not find a clear reason why models trained with CADec underperformed even our baseline, but one possible explanation could be that it uses three context sentences at once for each sentence and does not overlap them with the previous and next four-sentence lines, which effectively shrinks the training data down to $\frac{1}{4}$th of the original size.

For comparison, we also trained NMT models on WMT20 data ($\sim$13M parallel sentences, excluding \textit{News Commentary v15}; WMT column in Table \ref{tab:mt-results}). For these models, we used \textit{newsdev2020} as development data and \textit{News Commentary v15}\footnote{http://www.statmt.org/wmt20/translation-task.html} as evaluation data since \textit{newstest2020} was not yet available at the time and for Japanese \textit{News Commentary v15} was only 1811 sentences long. These models reached 21.14 BLEU for \enja \ and 20.43 BLEU for \jaen \ on \textit{News Commentary v15}, but on our evaluation data they under-performed our baselines. This shows that even with 60x the training data these models struggle to translate conversations. By combining all training data the gain over the baselines is only 0.81 - 1.46 BLEU.

\begin{CJK}{UTF8}{ipxm}
Figure \ref{fig:mt-comparison} shows one example of a Japanese sentence and its translations by the MT systems. There are no pronouns in the source sentence, but there is the noun 「方」, which should be translated into the English pronoun ``he", specifying the person to be the successor to the store. Both systems manage to translate this part correctly, but the baseline generates an additional pronoun in the end instead of ``the store". We observed many similar situations, where the contextual translation still didn't match the reference and was not perfect, but the selection of pronouns had improved.
\end{CJK}

\begin{table}[t]
    \centering
\begin{tabular}{|l|r|r|}
\hline
                & \textbf{Total}   & \textbf{Unique} \\ \hline
    Development & 2,051   & 2,012  \\ \hline
    Evaluation  & 2,120   & 2,070  \\ \hline
    Training    & 80,629  & 74,377 \\ \hline
    AMI         & 110,483 & 75,660 \\ \hline
    ON          & 28,429  & 24,335 \\ \hline
    \end{tabular}
    \caption{Total vs. unique sentence pairs of training, development and evaluation BSD data; and AMI and OntoNotes sub-corpora.}
    \label{tab:total-unique-other}
\end{table}

\begin{table}[t]
    \centering
\begin{tabular}{|l|c|c|}
\hline
                            & \textbf{\jaen}         & \textbf{\enja}                 \\ \hline
    WMT                     & 16.29         & 12.99                 \\ \hline
    WMT+                    & 18.44         & 15.33                 \\ \hline \hline 
    Baseline                & 16.98         & 14.52                 \\ \hline
    CADec                   & 15.31         & 12.55                 \\ \hline
    Concatenated            & 17.07         & 14.15                 \\ \hline
    Concatenated + factors  & 17.24         & 14.19                 \\ \hline
    \end{tabular}
    \caption{MT experiment results in BLEU scores. WMT uses only WMT 2020 data and WMT+ uses WMT 2020 along with our corpus for training. The rest use only our corpus for training.}
    \label{tab:mt-results}
\end{table}

\begin{CJK}{UTF8}{ipxm}
\begin{figure}[t]
    \begin{small}
    \begin{tabular}{@{}l@{}p{14cm}}
    
    \bf Source: & おっ、きっとお店の後継者になる方ですね。\\
    \bf Reference: & Oh, he must be the successor to the store.\\
    \bf Baseline: & Oh, I'm sure he will succeed \textbf{you}.\\
    \bf Con.+fact.: & Oh, I'm sure he will be the successor to the store.\\
    
    \end{tabular}
    \caption{\jaen \ translations of a sentence where the baseline generated an incorrect pronoun, but the concat. + factors system produced a more fitting translation.}
    
    \label{fig:mt-comparison}
    \end{small}
\end{figure}
\end{CJK}

\section{Human Evaluation}
\label{sec:human-eval}

We translated the evaluation set in both directions using our baseline NMT and performed a two step human evaluation similar to \newcite{voita-etal-2019-good}. After that, we analysed the remaining sentences to determine which truly require context.

We used Yahoo! Japan Crowdsourcing\footnote{https://crowdsourcing.yahoo.co.jp/} for the human evaluation. Evaluation quality was guaranteed using screening questions which were indistinguishable from the real questions. Only those who correctly answered all the screening questions were considered valid evaluators. Each sentence was evaluated by 5 different evaluators.

In the first step, evaluators were asked to mark each sentence individually as OK or Not Good (NG), where OK meant that the general meaning of the original sentence was transferred to the translation, whereas NG meant that the translation is completely unusable. In the second step, we used only the consecutive pairs of sentences, which were both marked as OK in the first step by at least three evaluators, and asked evaluators to mark them as OK if the corresponding translations made sense in context of each other. We calculated the Free-Marginal Kappa \cite{randolph2005free} values for the evaluations to measure agreement between evaluators. The results (overall agreement - 67\%, Free-marginal kappa - 0.34) show moderate agreement, which is common for crowdsourcing.

\subsection{Analysis}

As a result of the crowdsourcing campaign (Table \ref{tab:human-eval-table}) we had 228 \enja \ sentence pairs and 208 \jaen \ sentence pairs marked as NG in context of each other. We employed two linguistic experts to check the translations along with their respective sources and references to determine their ambiguity and need for additional context. For this step they were also asked to categorise the ambiguity type. 

After the final step 9 \enja \ and 43 \jaen \ sentence pairs were marked as context-dependent. 38 \jaen \ pairs lack pronouns in the source sentence and do not have enough content to produce an unequivocal translation. 
The other 5 \jaen \ pairs contain ambiguous words or phrases, which can be translated differently, depending on the context. For example, \begin{CJK}{UTF8}{ipxm}「1組」\end{CJK} can be translated as either ``one couple" or ``one group". Similarly in \enja , Chinese can refer to language (\begin{CJK}{UTF8}{ipxm}中国語\end{CJK}) or food (\begin{CJK}{UTF8}{ipxm}中華料理\end{CJK}) as shown in Figure \ref{fig:ambiguity-example}. Our best contextual models still struggle to translate such ambiguities, while slightly outperforming SL baselines in handling pronouns. 

Figure \ref{fig:ambiguity-example-2} shows example mistranslations of pronouns, where they are omitted (as is often done in the spoken language) on the Japanese side, but expected in the English translation. The contextual MT model does get some of the pronouns right in the first sentence, but perhaps requires longer context for the second one.

\begin{CJK}{UTF8}{ipxm}
\begin{figure}[t]
\begin{small}
  \begin{tabular}{@{}l@{ }l@{}}
    \bf Previous Source: & What kind of food should we choose?\\
    \bf Previous Reference: & どういうジャンルにしますか？\\
    \bf Previous MT: & どんな食べ物を選ぶべきか。\\ \hline

    \bf Source: & How about \textbf{Chinese}?\\
    \bf Reference: & \textbf{中華料理}はどう？\\
    \bf MT: & \textbf{中国語}はどうですか？\\

  \end{tabular}
  \caption{\enja \ MT output where {\it Chinese} is translated into ``中国語'' (Chinese language) instead of ``中華料理'' (Chinese food).}

  \label{fig:ambiguity-example}
  \end{small}
\end{figure}
\end{CJK}

\begin{table}[t]
    \centering
    \begin{tabular}{|c|c|c|c|c|c|}
    \hline
     \multicolumn{2}{|c|}{\textbf{EN$\rightarrow$RU}} & \multicolumn{2}{c|}{\textbf{EN$\rightarrow$JA}} & \multicolumn{2}{c|}{\textbf{JA$\rightarrow$EN}} \\ \hline
     \multicolumn{2}{|c|}{2000} & \multicolumn{2}{c|}{2051} & \multicolumn{2}{c|}{2051} \\ \hline
     NG & OK & NG & OK & NG & OK \\ \hline 
     140 & 1649 & 228 & 931 & 208 & 1174 \\ \hline
     4\% & 41\% & 11\% & 45\% & 10\% & 57\% \\ \hline
    \end{tabular}
    \caption{Results of the second step of the crowdsourcing human evaluation compared to EN$\rightarrow$RU \cite{voita-etal-2019-good}. The first row shows sentence pair totals and the last two rows show sentence pairs, where both sentences were marked as ``good" individually, evaluated in context of each other as either good or bad pairs.}
    \label{tab:human-eval-table}
\end{table}

\begin{CJK}{UTF8}{ipxm}
\begin{figure}[t]
\begin{small}
  \begin{tabular}{@{}l@{ }l@{}}
    \bf Prev. Source: & いつ 返事 くれる と 言っ て た？\\
    \bf Prev. Reference: & Did they say when they will get back to you?\\
    \bf Prev. Base.: & when did you say you' d answer me?\\
    \bf Prev. Conc.+f.: & When did they say they will reply?\\ \hline

    \bf Source: & 来週 早々 に は 、 と 言っ て まし た 。\\
    \bf Reference: & They said early next week.\\
    \bf Base.: & He told me early next week.\\
    \bf Conc.+f.: & I said it early next week .\\

  \end{tabular}
  \caption{\jaen \ MT output by baseline (Base.) and concatenated context + factored (Conc.+f.) models of sentences with no pronouns in the source and expected pronouns in the translation.}
  \label{fig:ambiguity-example-2}
  \end{small}
\end{figure}
\end{CJK}

\section{Conclusion}
\label{sec:conclusion}

We presented a document-aligned parallel corpus of English-Japanese conversations intended for training and evaluation of MT systems. We describe the corpus in detail and indicate which linguistic phenomena are challenging for MT. In our evaluation set we marked examples, which can have multiple contrasting translations when tackled on the sentence-level. The release will include the full BSD corpus and Japanese translations of AMI and ON along with instructions on how to align them. The original source language, speaker, scene, document, ambiguity type will also be included. 

In the future we plan to model speakers and origin languages in MT, as it can help capture broader context \cite{maruf-etal-2018-contextual} and more precise pronoun translations \cite{vanmassenhove-etal-2018-getting}. We are also interested in experimenting with modelling the scene information within the training data to produce more appropriate translations for each of the politeness settings.

\section*{Acknowledgements}

This work was supported by “Research and Development of Deep Learning Technology for Advanced Multilingual Speech Translation”, the Commissioned Research of National Institute of Information and Communications Technology (NICT), JAPAN.

\bibliography{anthology,other}
\bibliographystyle{EMNLP/acl_natbib}


\end{document}